\documentclass[conference,compsoc]{IEEEtran}
\usepackage[normalem]{ulem}
\usepackage{multirow}

\ifCLASSOPTIONcompsoc
\usepackage[nocompress]{cite}
\else
\usepackage{cite}
\fi
\ifCLASSINFOpdf
\usepackage[pdftex]{graphicx}
\else
\fi
\usepackage{amsmath}
\usepackage{amssymb}
\begin{document}
	\title{Joint RNN Model for Argument Component Boundary Detection}
	
	\author{\IEEEauthorblockN{Minglan Li, Yang Gao, Hui Wen, Yang Du, Haijing Liu and Hao Wang}
		\IEEEauthorblockA{Institute of Software Chinese Academy of Sciences\\
			University of Chinese Academy of Sciences\\
			Email: \{minglan2015, gaoyang, wenhui2015, duyang2015, haijing2015, wanghao\}@iscas.ac.cn}}
	
	
	% compsoc conferences compared to traditional conferences), use this
	%Telephone: (800) 555--1212, Fax: (888) 555--1212}


	\maketitle
	
	\begin{abstract}
		\emph{Argument Component Boundary Detection} (ACBD) is an important sub-task 
		in argumentation mining; it aims at identifying the word sequences 
		that constitute argument components, and is usually considered 
		as the first sub-task in the argumentation mining pipeline. 
		Existing ACBD methods heavily depend on task-specific knowledge,
		and require considerable human efforts on feature-engineering.
        To tackle these problems, in this work, we formulate ACBD as a sequence labeling problem
		and propose a variety of \emph{Recurrent Neural Network} (RNN) based 
		methods, which do not use domain specific or 
		handcrafted features beyond the relative position 
		of the sentence in the document. 
		In particular, we propose a novel \emph{joint RNN model} 
		that can predict whether sentences are argumentative or not, 
		and use the predicted results to more precisely 
		detect the argument component boundaries. 
		We evaluate our techniques on two corpora from two different genres;
		results suggest that our joint RNN model obtain 
		the state-of-the-art performance on both datasets.
	\end{abstract}
	% \fi
	\IEEEpeerreviewmaketitle
	
	\section{Introduction}
	
	\emph{Argumentation mining} aims at automatically extracting 
	\emph{arguments} from natural language texts \cite{moens2013argumentation}.
	An \emph{argument} is a basic unit people use to 
	persuade their audiences to accept
	a particular state of affairs \cite{eckle2015role},
	and it usually consists of one or more
	\emph{argument components}, for example
	a \emph{claim} and some \emph{premises} offered in support of the claim.
	As a concrete example, consider the essay excerpt below
	(obtained from the essay corpus in \cite{stab2016parsing}):
	\begin{quote}
		\textbf{Example 1}: 
		Furthermore, \textcircled{1}[[\uline{putting taxpayers' money on 
			building theaters or sports stadiums is unfair to those who cannot use them}]]. 
		That is the reason why \textcircled{2}[[\textbf{sectors such 
			as medical care and education deserve more governmental support}]], 
		because \textcircled{3}[[\uline{they are accessed by every individual 
			in our society on a daily basis}]].
	\end{quote}
	
	The above example includes three argument components 
	([[ ]] give their boundaries): one claim (in bold face) and two premises (underlined). 
	Premises \textcircled{1} and \textcircled{3} support the claim \textcircled{2}.
	As argumentation mining reveals the discourse relations between 
	clauses, it can be potentially used in applications like 
	decision making, document summarising, essay scoring, etc.,
	and thus receives growing research interests in recent years
	(see, e.g. \cite{lippi2016argumentation}).
	
	A typical argumentation mining pipeline consists of 
	three consecutive subtasks\cite{stab2014identifying}:
	i) separating argument components from non-argumentative texts, 
	ii) classifying the type (e.g. claim or premise or others)
	of argument components; and 
	iii) predicting the relations (e.g. support or attack)
	between argument components. 
	The first subtask is also known as
	\emph{argument component boundary detection} (ACBD);
	it aims at finding the exact boundary of a consecutive token subsequence 
	that constitutes an argument component, thus separating it from non-argumentative texts.
	In this work, we focus on the ACBD subtask, because
	ACBD's performance significantly influences
	downstream argumentation mining subtasks' performances,
	but there exist relatively little research working on ACBD.
	
	Most existing ACBD techniques require
	sophisticated hand-crafted features 
	(e.g. syntactic, structural and lexical features) 
	and domain-specific resources (e.g. indicator gazetteers),
	resulting in their poor cross-domain applicabilities.
	To combat these problems,
	in this work, we consider ACBD as a sequence labeling task at the token level and
	propose some novel neural network based ACBD methods,
	so that no domain specific or hand-crafted features
	beyond the relative location of sentences are used.
	Although neural network based approaches have 
	been recently used in some related Natural Language Processing
	(NLP) tasks,
	such as linguistic sequence labelling\cite{huang2015bidirectional} and 
	named entity recognition (NER)\cite{lample2016neural}, applying neural network to 
	ACBD is challenging because the length of
	an argument component is much longer than that of a
	name/location in NER: 
	\cite{stab2014annotating} reports that an argument component
	includes 24.25 words in average, while a name/location usually
	consists of only 2 to 5 words.
	In fact, it has been reported in \cite{stab2016parsing,levy2014context} that 
	separating argumentative and non-argumentative texts 
	is often subtle even to human annotators.
	
	In particular, our neural network models are designed to capture two intuitions. 
	First, since an argument component often consists of 
	considerable number of words, it is essential to
	jointly considering multiple words' labels
	so as to detect argument components' boundaries; hence,
	we propose a bidirectional \emph{Recurrent Neural Network}
	(RNN)  \cite{goller1996learning} with a \emph{Conditional Random Field} (CRF) 
	\cite{lafferty2001conditional} layer above it, 
	as both RNN and CRF are widely recognized as
	effective methods for considering contextual information.
	Second, we believe that
	if the argumentative-or-not information of each 
	sentence\footnote{
		In this work, we assume that an argument component cannot 
		span across multiple sentences. This assumption is valid
		in most existing argumentation corpora, e.g. \cite{stab2016parsing}.}
	is provided a priori, the performance of ACBD can be 
	substantially improved. 
	As such, we propose a novel \emph{joint RNN model} that can 
	predict a sentence's argumentative status and use
	the predicted status to detect boundaries.
	
	The contributions of this work are threefold: 
	i) we present the first deep-learning based ACBD technique, 
	so that the feature-engineering demand is greatly reduced  
	and the technique's cross-domain applicability is significantly improved;
	ii) we propose a novel joint RNN model that can classify
	the argumentative status of sentences and 
	separating argument components from non-argumentative texts 
	simultaneously, which can significantly improve
	the performance of ACBD;
	and iii) we test our ACBD methods on two different
	text genres, and results suggest that our approach 
	outperforms the state-of-the-art techniques in both domains.

	\section{Related Work}
	
	In this section, we first review ACBD techniques, and then 
	review works that apply RNN to applications related to
	ACBD, e.g. sequence labeling and text classification.
	
	\subsection{Boundary Detection}
	\label{subsec:back:acbd}
	Most existing ACBD methods consist of two consecutive subtasks: 
	identifying argumentative sentences (i.e. sentences that include
	some argument components)
	and detecting the component boundaries \cite{lippi2016argumentation}. 
	Levy et al. \cite{levy2014context} identify context-dependent claims in 
	Wikipedia articles by using a cascade of classifiers. 
	They first use logistic regression to identify sentences containing 
	topic-related claims (the topic is provide a priori), and
	then detect the boundaries of 
	claims and rank the candidate boundaries, so as to identify the most relevant claims for 
	the topic. However, the importance of topic information is questionable, 
	as Lippi and Torroni \cite{lippi2015context} achieve a similar 
	result on the first subtask without using the topic information. 
	Goudas et al. \cite{goudas2014argument} propose a ACBD
	technique and test it on a corpus constructed from social media texts. 
	They first use a variety of classifiers 
	%including Support Vector Machines, Naïve Bayes, Random Forest and Logistic Regression 
	to perform the first subtask, and then employ a feature-rich CRF to perform
	the second subtask. 
	%Sardianos et al. \cite{sardianos2015argument} extend
	%Goudas et al.'s work by using additional features like distributed 
	%representations of words \cite{mikolov2013distributed} and 
	%a small gazetteer containing a list of cue words.
	
	Besides the two-stage model presented above, 
	some works consider ACBD as a sequence labeling task at token level.
	Stab and Gurevych \cite{stab2016parsing} employ a CRF model 
	with four kinds of hand-craft features (structural, syntactic, lexical and 
	probability features) to perform ACBD on their persuasive essay corpus.
	Unlike texts in Wikipedia, persuasive essays are organised by 
	structurally and the percentage of argumentative sentences
	are much higher (77.1\% sentences in persuasive essays include
	argument component). The performance of this ACBD technique
	(in terms of macro F1) is .867.
	
	\subsection{RNN on Similar Tasks}
	\label{subsec:back:rnn}
	RNN techniques, especially Long Short-Term Memory (LSTM)
	\cite{hochreiter1997long}, have recently been successfully applied 
	to sequence labeling and text classification tasks in various NLP problems. 
	%Here we mainly review the applications of RNN
	%on sequential labeling tasks, because these tasks are similar 
	%to ACBD.
	Graves et al. \cite{graves2013speech} propose a bidirectional RNN for
	speech recognition, which takes the context on both sides of each word into account.
	However, in sequential labeling tasks with strong dependencies 
	between output labels, the performance of RNN is not ideal.
	To tackle this problem, 
	instead of modeling tagging decisions independently, 
	Huang et al.\cite{huang2015bidirectional} and Lample et al. 
	\cite{lample2016neural} apply a sequential CRF to jointly decode labels for the whole sequence. 
	
	RNN has also be successfully used in text classification. 
	Lai et al. \cite{lai2015recurrent} propose a \emph{Recurrent Convolutional Neural Network},
	which augments a max-pooling layer after bidirectional RNN. 
	The purpose of the pooling layer is to capture the most 
	important latent semantic factors in the document. 
	Then the softmax function is used to predict 
    the classification distribution. 
    Results shows the effectiveness on the text classification tasks.
    %(YG: use one sentence to summarise the results).
	
	Some RNN-based techniques are developed for the spoken language 
	understanding task, in which both text classification and sequence labeling
    are involved: \emph{intent detection} is a classification problem,
    while \emph{slot filling} is a sequence labeling problem. Liu and Lane 
	\cite{liu2016attention} propose an attention-based bidirectional 
    RNN model to perform these two tasks simultaneously; the method
    achieves state-of-the-art performance on both tasks.
	
	\section{Models}
	\label{sec:models}
	We consider a sentence in the document as a sequence of 
	tokens/words and label the argument boundaries using the IOB-tagset:
	a word is labelled as ``B'' if it is the first token in 
	an argument component, ``I'' if it constitutes, but not leads,
	an argument component, and ``O'' if it is not
	included in any argument components. 
	In this section, we first review some widely used
	techniques for sequence labeling
	, 
	%:
	%Bidirectional-LSTM (Bi-LSTM) in  
	%Sect. \ref{subsec:models:bi-lstm}, CRF in \ref{subsec:models:crf}, 
	%Bi-LSTM-CRF in Sect. \ref{subsec:models:lstm-crf},
	%and attention-based RNN for text classification in Sect. 
	%\ref{subsec:models:attention-rnn}; 
	then we present our joint RNN model, 
	which can distinguish argumentative and non-argumentative
	sentences and use this information to detect boundaries, in 
	Sect. \ref{subsec:models:joint-rnn}.
	
	\subsection{Bi-LSTM}
	\label{subsec:models:bi-lstm}
	RNN \cite{goller1996learning} is a neural architecture 
	designed for dealing with sequential data. 
	%In RNN, the current output depends on the previous states, 
	%which enables RNN to remember history information. 
	RNN takes as input a vector 
	$\textbf{X}=[\textbf{x}_{t}]_{1}^{T}$    and returns a feature vector sequence 
	$\vec{\textbf{h}}= [\textbf{h}_{t}]_{1}^{T}$
	at every time step.\footnote{In this work,
		we let $[x]_1^T$ be the short-hand notation for vector
		$[x_1, \cdots, x_T]$, where $T \in \mathbb{N}^*$ is the length
		of the vector.}
	A primary goal of RNN is to capture long-distance dependencies.
	However, in many real applications, standard RNN are often 
	biased towards their most recent inputs in the sequence 
	\cite{bengio1994learning}
	%, fail to effectively ``memorise'' historical information
	. 
	LSTM \cite{hochreiter1997long} alleviates this problem 
	by using a memory cell and three gates 
	(an input gate $\textbf{i}$, a forget gate $\textbf{f}$, and an output gate $\textbf{o}$) 
	to tradeoff between the influence of the new input state and 
	the previous state on the memory cell. 
	The computation operations of a memory cell $\textbf{c}_{t}$ and 
	hidden states $\textbf{h}_{t}$ (of size $H$),
	at time step $t$, are as follows: 
	\begin{equation*}
	\begin{split}
	&\textbf{i}_{t}=\sigma(\textbf{W}^{i}\textbf{x}_{t}+\textbf{U}^{i}\textbf{h}_{t-1}+\textbf{b}^{i}),\\
	&\textbf{f}_{t}=\sigma(\textbf{W}^{f}\textbf{x}_{t}+\textbf{U}^{f}\textbf{h}_{t-1}+\textbf{b}^{f}),\\
	&\textbf{o}_{t}=\sigma(\textbf{W}^{o}\textbf{x}_{t}+\textbf{U}^{o}\textbf{h}_{t-1}+\textbf{b}^{o}),\\
	&\textbf{g}_{t}=tanh(\textbf{W}^{g}\textbf{x}_{t}+\textbf{U}^{g}\textbf{h}_{t-1}+\textbf{b}^{g}),\\
	&\textbf{c}_{t}=\textbf{f}_{t}\odot \textbf{c}_{t-1}+\textbf{i}_{t}\odot \textbf{g}_{t},\\
	&\textbf{h}_{t}=\textbf{o}_t\odot tanh(\textbf{c}_{t}),\\
	\end{split}
	\end{equation*}
	
	\noindent
	where
	%$\textbf{x}_t$ is the input vector at time step $t$, 
	$\textbf{b}^y$ is the bias vector for gate $y$
	(where $y$ can be $i$, $f$, $o$ or $g$), 
	$\sigma$ is the element-wise sigmoid function, 
	and $\odot$ is the element-wise multiplication operator. 
	$\textbf{W}\in R^{H\times d} $, 
	$\textbf{U} \in R^{H\times H}$ and $\textbf{b}\in R^{H\times 1}$ 
	are the network parameters.
	
	The LSTM presented above is known as \emph{single direction LSTM},
	because it only considers the preceding states, ignoring the
	states following the current state; thus,  it fails to 
	consider the ``future'' information.
	Bidirectional LSTM (Bi-LSTM) \cite{graves2005framewise} is proposed to 
	combat this problem. 
	Bi-LSTM includes a forward LSTM 
	and a backward LSTM, thus can capture both past and future 
	information. 
	Then the final output of Bi-LSTM is the product of
	concatenating the past and future context representations:
	$\textbf{h}_t=[\overrightarrow{\textbf{h}_t};\overleftarrow{\textbf{h}_t}]$,
	where $\overrightarrow{\textbf{h}_t}$ and $\overleftarrow{\textbf{h}_t}$
	are the forward and backward LSTM, resp.
	
	\subsection{CRF}
	\label{subsec:models:crf}
	CRF \cite{lafferty2001conditional} is widely used in sequence labeling tasks.
	For a given sequence $\textbf{X}$ and its labels 
	$\textbf{y}$, CRF gives a real-valued score as follows:
	\begin{equation*}
	score(\textbf{X}, \textbf{y})=\sum_{t=2}^{T}\psi(\textbf{y}_{t-1}, \textbf{y}_{t}) + \sum_{t=1}^{T}\phi(\textbf{y}_{t}),
	\end{equation*}
	\noindent where $\phi(\textbf{y}_{t})$ is the unary potential 
	for the label at position $t$ and $\psi(\textbf{y}_{t-1}, \textbf{y}_{t})$ 
	is the pairwise potential of labels at $t$ and $t-1$. 
	The probability of \textbf{y} given \textbf{X} can
	be obtained from the score:
	\begin{equation}
	\label{eq:training}
	p(\textbf{y}|\textbf{X})=\frac{1}{Z}exp(score(\textbf{X}, \textbf{y})).
	\end{equation}
	Given a new input $\textbf{X}_{new}$,
	the goal of CRF is find a label $\textbf{y}^*$ for $\textbf{X}_{new}$,
	whose conditional probability is maximised:
	\begin{equation}
	\label{eq:decoding}
	\textbf{y}^*=\arg\max_{\textbf{y}}(\sum_i log(p(\textbf{y}|\textbf{X}_{new}))).
	\end{equation}
	The process for obtaining the optimal label is termed \emph{decoding}.
	For a linear chain CRF described above 
	that only models bigram interactions between outputs, 
	both training (Eq. \eqref{eq:training}) and decoding (Eq. \eqref{eq:decoding}) 
	can be solved efficiently by dynamic programming.
	
	\subsection{Bi-LSTM-CRF}
	\label{subsec:models:lstm-crf}
\begin{figure}[!t]
		\centering
		\includegraphics[width=2.8in]{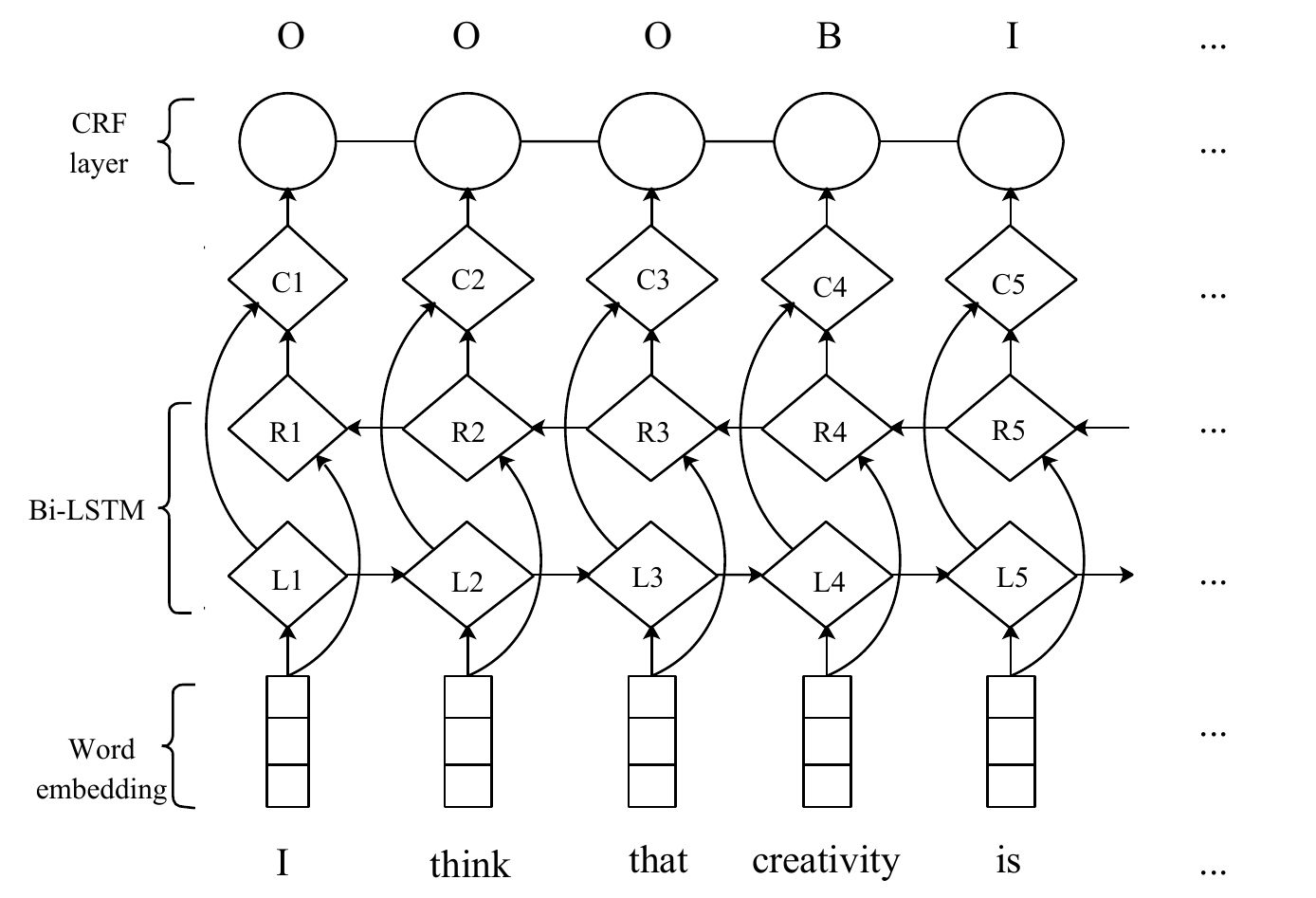}
		\caption{The structure of Bi-LSTM-CRF network.
			\label{fig:bi-lstm-crf}}
	\end{figure}

	In sequential labeling tasks where there exist strong dependencies 
	between neighbouring labels, the performance of Bi-LSTM is not ideal.
	To tackle this problem, Huang et al. \cite{huang2015bidirectional}
	propose the Bi-LSTM-CRF method, which augments a CRF layer
	after the output of Bi-LSTM, so as to explicitly model the 
	dependencies between the output labels. Fig. \ref{fig:bi-lstm-crf}
	illustrates the structure of Bi-LSTM-CRF networks.
	
	For a given input sentence $\textbf{X}=[\textbf{x}_{t}]_{1}^{T}$, 
	$\textbf{h}= [\overrightarrow{\textbf{h}_L};
	\overleftarrow{\textbf{h}_R}]$ is the output of Bi-LSTM, where 
	$\overrightarrow{\textbf{h}_L}$ and $\overleftarrow{\textbf{h}_R}$ are 
	the output of the forward and backward LSTM, resp. 
	%(see Sect. \ref{subsec:models:bi-lstm} for Bi-LSTM). 
	The connection layer $C$ is used for connecting the 
	\emph{structure features} $s$ and the output of Bi-LSTM,
	namely the feature representations $\textbf{h}$.
	Note that $s$ is the relative position of the 
	input sentence in document, and is not shown in Fig. \ref{fig:bi-lstm-crf}. 
	The output of $C$ is a matrix of scores, denoted by $\textbf{P}$.
	$\textbf{P}$ is of size $T \times k$, where $k$ is the number 
	of distinct tags, and $\textbf{P}_{ij}$ corresponds to the 
	score of the $j^{th}$ tag of $i^{th}$ word in a sentence. 
	The score $s$ of a sentence $\textbf{X} = [\textbf{x}_{t}]_{1}^{T}$ along with 
	a path of tags $\textbf{y} = [y_i]_{1}^{T}$ is then defined as follows:
	\begin{equation*}
	score(\textbf{X}, \textbf{y})=\sum_{i=0}^{T}\textbf{A}_{y_{i},y_{i+1}} + \sum_{i=1}^{T}\textbf{P}_{i,y_i},
	\end{equation*}
	where  $\textbf{A}$ is the \emph{transition matrix}, which
	gives the transition scores between tags such that 
	$\textbf{A}_{i,j}$ is the score of a transition 
	from the tag $i$ to tag $j$. We add two special tags at the beginning 
	and end of the sequence so that $\textbf{A}$ is a squared matrix 
	of size $k+2$. The conditional probability for a label sequence 
	\textbf{y} given a sentence \textbf{X} thus 
	can be obtained as follows:
	\begin{equation*}
	p(\textbf{y}|\textbf{X})=\frac{e^{score(\textbf{X}, \textbf{y})}}{\sum_{\widetilde{\textbf{y}} \in \textbf{Y}_{\textbf{X}}} e^{score(\textbf{X},\widetilde{\textbf{y}})}},
	\end{equation*}
	where $\textbf{Y}_{\textbf{X}}$ represents all possible tag sequences 
	for a input sentence $\textbf{X}$. The network is trained by 
	minimizing the negative log-probability of the correct 
	tag sequence \textbf{y}.
	%\begin{equation*}
	%-log(p(\textbf{y}|\textbf{X}))=log(\sum_{\widetilde{\textbf{y}} \in %{\textbf{Y}_{\textbf{X}}}} e^{score(\textbf{X},\widetilde{\textbf{y}})}) - %score(\textbf{X}, \textbf{y}).
	%\end{equation*}
	Dynamic programming techniques can be used to efficiently 
	compute the transition matrix $A$ and the optimal tag 
	sequence $\textbf{y}^*$ for inference.
	
	\subsection{Attention based RNN for Classification}
	\label{subsec:models:attention-rnn}
	\begin{figure}[!t]
		\centering
		\includegraphics[width=2.5in]{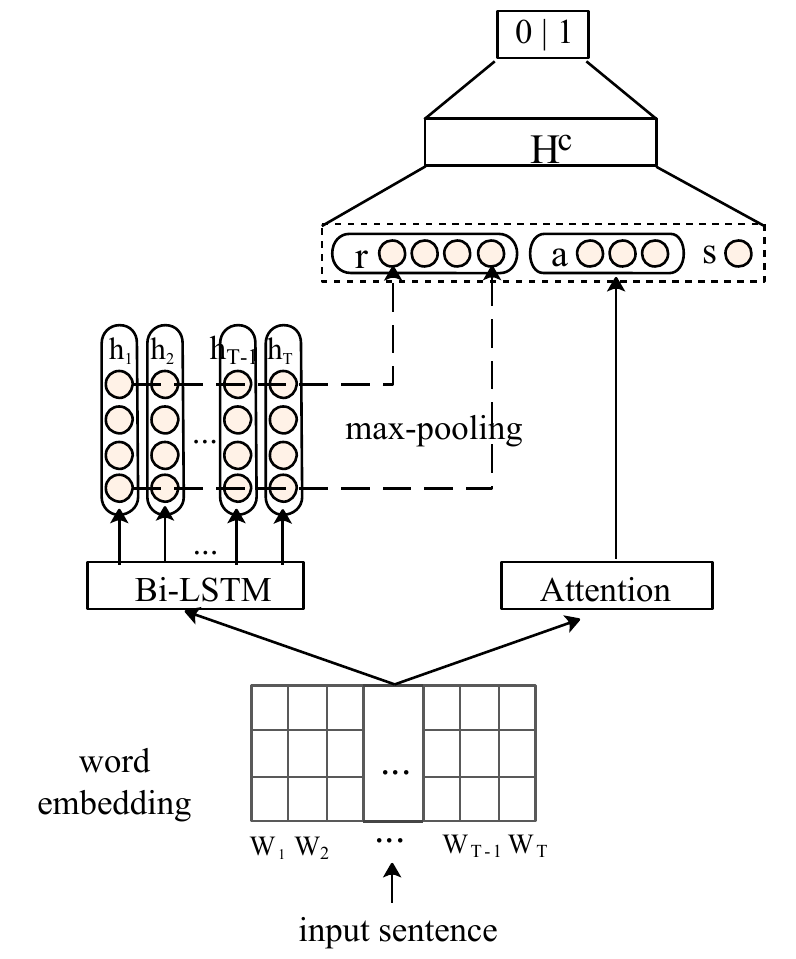}
		\caption{an attention-based RNN network for classification
			\label{fig:attention-rnn}}
	\end{figure}
	
	Besides in sequence labeling, RNN is also widely used in text classification
	tasks. Lai et al. \cite{lai2015recurrent} combine the word embeddings and representation output by Bi-LSTM as the feature representation for text classification, weighting each input word equally.
	However, as the importances of words differ, such a equal-weighting
	strategy fails to highlight the truly important information.
	The \emph{attention} mechanism \cite{rush2015neural}
	is proposed to tackle this problem. As the name suggests,
	the attention mechanism computes a weight vector to measure 
	the importance of each word and aggregates those informative 
	words to form a sentence vector. Specifically,
	\begin{align*}
	& f(\textbf{x}_t)=tanh(\textbf{W}^a \textbf{x}_t+\textbf{b}^a),  \\
	& \alpha_t=\frac{exp(\textbf{U}^af(\textbf{x}_t))}{\sum_i exp(\textbf{U}^af(\textbf{x}_i))}, \\
	& \textbf{a} = \sum_t\alpha_t \cdot \textbf{x}_t. 
	\end{align*}
	The sentence vector $\textbf{a}$ is the weighted sum of 
	the word embeddings $\textbf{x}_t$, weighted by $\alpha_t$. 
	Vector $\textbf{a}$ gives additional supporting information, 
	especially the information that requires longer term dependencies;
	this information can hardly be fully captured by the hidden states.
	
	The architecture of the RNN for classification is illustrated 
	in Fig \ref{fig:attention-rnn}.
	Remind that, Bi-LSTM  can capture both the past and the future context 
	information and can convert the tokens comprising each document 
	into a feature representation $\textbf{h}=[\textbf{h}_t]_1^T$. 
	Max-pooling operation is used to extract 
	maximum values over the time-step dimensions of 
	$\textbf{h}$ to obtain a sentence level representation $\textbf{r}$.
	%(YG: where does the superscript c comes from? What is $h^c$??
	%Your description below does not explain your formula below)
	Then the sentence's argumentative status is predicted 
	by the concatenating of context feature $\textbf{r}$, 
	weighted sentence vector $\textbf{a}$ and 
    structure feature $s$ (relative location of the sentence):
	\begin{equation*}
	\textbf{H}^c = softmax({\textbf{W}}^{c} [\textbf{r};\textbf{a};s] + \textbf{b}^c)
	\end{equation*}
	where $\textbf{H}^c$ is the output of softmax function, which represents the probabilities of sentence's argumentative status.
	
	\subsection{Joint RNN Model}
	\label{subsec:models:joint-rnn}
	\begin{figure}[!t]
		\centering
		\includegraphics[width=2.35in]{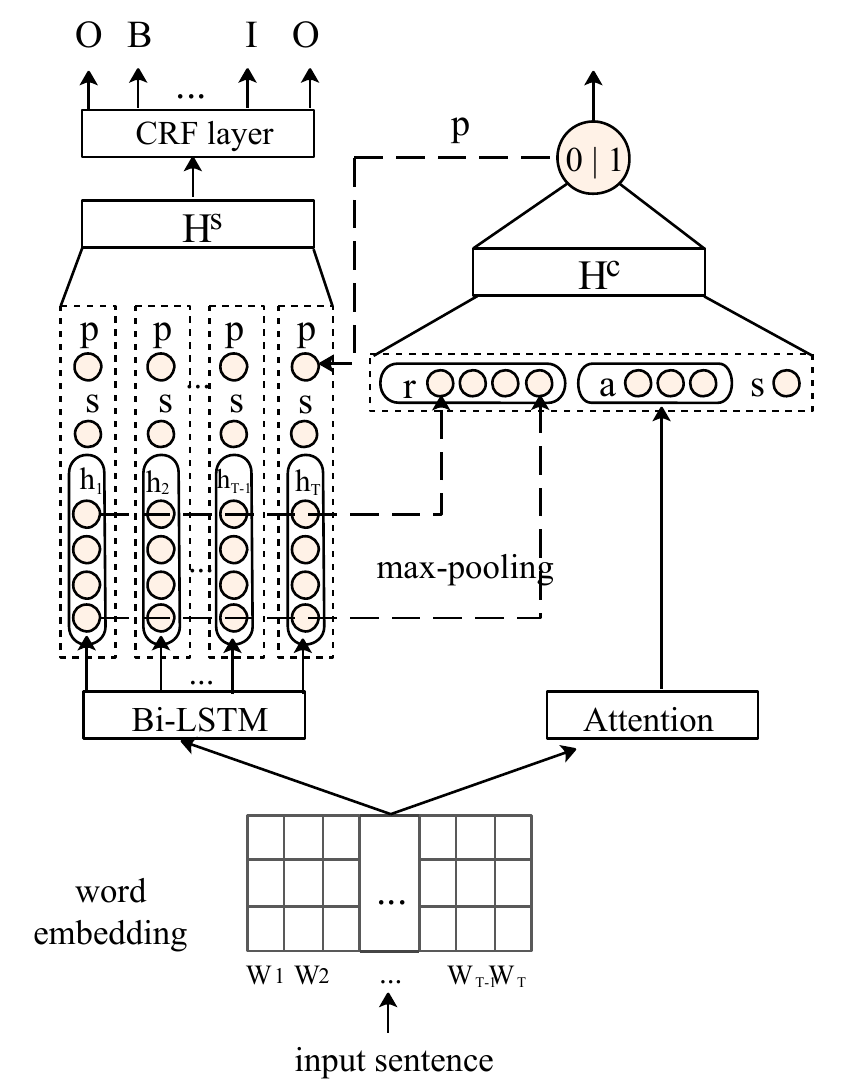}
		\caption{joint RNN network for classification and boundary detection
			\label{fig:joint-rnn}}
	\end{figure}
	The joint model for argumentative sentence classification and sequence labeling in boundary detection is shown in Fig \ref{fig:joint-rnn}. In the proposed model, a Bi-LSTM reads the source sentence in both forward and backward directions and creates the hidden states $\textbf{h}=[\textbf{h}_t]_1^T$. 
	For sentence argumentative classification, as we mentioned in Sect. \ref{subsec:models:attention-rnn}, an attention mechanism aggregate the input words into a sentence vector $\textbf{a}$. The max-pooling operation is applied to capture the key components of the latent information. the sentence's argumentative status $p$ is then predicted by the combination of vector $\textbf{a}$, vector $\textbf{r}$ (output by max-pooling operation) and relative location feature $s$. 
	
	For sequence labeling in boundary detection, we reuse the pre-computed hidden states $\textbf{h}$ of the Bi-LSTM. At each time-step, we combine each hidden state $\textbf{h}_t$  with the relative location feature $s$ and the sentence's predicted argumentative status $p$ created by the above mentioned classification operation:
	${\textbf{h}^{\\'}}_t = [\textbf{h}_t;s;p]$.
	Then the $\textbf{H}^{s}$ will be the scores matrix $\textbf{P}$ described in Sec. \ref{subsec:models:lstm-crf} which will be given to the CRF layer.
	\begin{equation*}
	{\textbf{H}^{s}}_t = tanh({\textbf{W}}^s \textbf{h}^{\\'}_t + \textbf{b}^s)
	\end{equation*}
	 The sequence labeling operation is as same as Sect. \ref{subsec:models:lstm-crf}. The network of joint model is trained to find the parameters that minimize the cross-entropy of the predicted and true argumentative status for sentence and the negative log-probability of the sentence's labels jointly.
	
	\section{Experiments}
	We first present the argumentation corpora on which 
	we test our techniques in Sect. \ref{subsec:exp:data},
	introduce our experimental settings in 
	Sect. \ref{subsec:exp:setting}, and present
	and analyse the empirical results in Sect. \ref{subsec:exp:result}.
	
	\subsection{Datasets}
	\label{subsec:exp:data}

	We evaluate the neural network based ACBD techniques 
	on two different corpora: the persuasive essays corpus 
	\cite{stab2016parsing} and the Wikipedia corpus 
    \cite{aharoni2014benchmark}.
	The persuasive essay corpus has three types of argument components:
	\emph{major claims}, \emph{claims} and \emph{premises}. 
	The corpus contains 402 English essays on a variety of topics,
	consisting of 7116 sentences and 147271 tokens (words).
	%LML the Wikipedia corpus also have envidence components, but in order to compare with levy's work, we only focus on claims. So i change the description here
	The Wikipedia corpus contains 315 Wikipedia articles grouped into 33 topics, 
    and 1392 \emph{context-dependent claims} have been annotated in total. 
	A context-dependent claim is \emph{
		``a general, concise statement that directly supports or contests the given topic''},
	thus claims that do not support/attack the claim are not annotated. 
	Note that the Wikipedia corpus is very imbalanced: 
    only 2\% sentences are argumentative (i.e. contain some argument components).
    %YG: why the same corpus has two cites?
    %it was reported in \cite{levy2014context} that only around $2\%$ sentences
	%are argumentative. 
	
	\subsection{Experiment Settings}
	\label{subsec:exp:setting}
	%LML: two corpus have different evaluation method
	On the persuasive essay corpus, in line with  
    \cite{stab2016parsing}, we use precision (P),
    recall (R) and macro-F1 as evaluation metrics,
    and use the same train/test split ratio: 322 essays are 
    used in training, and the remaining 80 essays are used in test.
	On the Wikipedia corpus, in line with \cite{levy2014context},
	a predicted claim is considered as True Positive if and only 
	if it precisely matches a labeled claim. 
	For all the articles across 33 topics, we randomly select 
    1/33 of all the sentences to serve as the test set, and 
    the remaining sentences are used in training. As the corpus is very imbalanced,
	we apply a random under-sampling on the training set
	so ensure that the ratio between non-argumentative and 
	argumentative sentences is 4:1. 
	
	In experiments on both corpora, we randomly select 10\% data in
    the training set to serve as the validation set. Training occurs for 200 epochs.
    Only the models that perform best (in terms of F1)
    on the validation set are tested on the test set.
    The RNN-based methods read in texts sentence by sentence,
    and each sentence is represented by concatenating 
    all its component words' embeddings.
    All RNN are trained using the Adam update rule
    \cite{kingma2014adam} with initial learning rate 0.001. 
    We let the batch size be 50 and the attention hidden size be 150. 
    To mitigate overfitting, we apply both the dropout method 
    \cite{srivastava2014dropout} and L2 weight regularization.
    We let dropout rate be 0.5 throughout our experiments 
    for all dropout layers 
    on both the input and the output vectors of Bi-LSTM. 
    The regularization weight of the parameters is 0.001. 
    Some hyper-parameter settings of the RNNs may 
    depend on the dataset being used. 
    In experiments on persuasive essays, we use Google's 
    word2vec \cite{mikolov2013distributed} 300-dimensional as 
    the pre-trained word embeddings, and set the hidden size to 150 
    and use only one hidden layer in LSTM;
	on Wikipedia articles, we use glove's 300-dimensional embeddings 
    \cite{pennington2014glove}, and let the hidden size of LSTM be 80.

	\subsection{Results and Discussion}
	\label{subsec:exp:result}
	\begin{table}[!t]
		\footnotesize
		\renewcommand{\arraystretch}{1.1} 
		\newcommand{\tabincell}[2]{\begin{tabular}{@{}#1@{}}#2\end{tabular}}
		\centering
		\caption{Boundary detection on persuasive essays.
			\label{table:essay}}
		\scalebox{0.9}{
			\begin{tabular}{|c|c|c|c|c|c|c|}
				\hline
				& F1 & P & R & F1-B & F1-I & F1-O \\
				\hline
				CRF (stab et al.\cite{stab2016parsing}) & 0.867 & 0.873 & 0.861 & 0.809 & 0.934 & 0.857 \\
				\hline
				
				\hline
				Human Upper Bound & 0.886 & 0.887 & 0.885 & 0.821 & 0.941 & 0.892 \\
				\hline
				\tabincell{c}{Bi-LSTM-CRF\\(with knowing \\argumentative status)}& 0.943 & 0.955 & 0.932 & 0.899 & 0.976 & 0.9546 \\
				\hline
				
				\hline
				Bi-LSTM & 0.825 & 0.850 & 0.806 & 0.775 & 0.910 & 0.791 \\
				\hline
				Bi-LSTM-CRF & 0.860 & 0.868 & 0.853 & 0.832 & 0.919 & 0.828 \\
				\hline
				\tabincell{c}{Joint RNN Model} & \textbf{0.873} & 0.893 & 0.857 & 0.839 & 0.931 & 0.848 \\
				\hline	
		\end{tabular}}
		\vspace{-6pt}
	\end{table}
	
	\begin{table}[!t]
		\footnotesize
		\renewcommand{\arraystretch}{1.1} 
		\centering
		\caption{Classification Results on Wikipedia Articles
			\label{table:classification}}
		\scalebox{0.9}{
			\begin{tabular}{|c|c|c|c|}
				\hline
				& Precison@200 & Recall@200 & F1@200 \\
				\hline
				Levy et al. \cite{levy2014context} & 0.09 & 0.73 & 0.16 \\
				%\hline
				TK \cite{lippi2015context} & 0.098 & 0.587 & 0.168 \\
				%\hline
				TK + Topic \cite{lippi2015context} & 0.105 & 0.629 & 0.180 \\
				\hline
				\hline
				& Precision & Recall & F1 \\
				\hline
				Joint RNN Model & 0.156 & 0.630 & 0.250 \\
				\hline
		\end{tabular}}
		\vspace{-6pt}
	\end{table}
	
	\begin{table}[!t]
		\footnotesize
		\renewcommand{\arraystretch}{1.1} 
		\centering
		\caption{Boundary Detection on Wikipedia Articles
			\label{table:boundary}}
		\scalebox{0.9}{
			\begin{tabular}{|c|c|c|c|}
				\hline
				& Precison@50 & Precison@20 & Precison@10 \\
				\hline
				Levy et al. \cite{levy2014context} & 0.120 & 0.160 & 0.200 \\
				\hline
				\hline
				& F1 & Precision & Recall \\
				\hline
				Bi-LSTM & 0.043 & 0.125 & 0.026 \\
				Bi-LSTM-CRF & 0.142 & 0.100 & 0.242 \\
				Joint RNN Model & 0.190 & 0.122 & 0.435\\
				\hline
		\end{tabular}}
	\end{table}
	
	%We first report experiments on the persuasive essays dataset. 
	%Our goal is to detect the exact boundary of the argument components, i.e. 
	%major claims, claims and premises. 
	The performance of different methods on the persuasive essays are presented in 
    Table \ref{table:essay}. Note that the performance of CRF is obtained
	from \cite{stab2016parsing}. Bi-LSTM achieves .825 macro F1 
	thanks to the context information captured by the LSTM layer. 
	Adding a CRF layer to Bi-LSTM can significantly 
	improve the performance and can achieve comparable results with 
	the CRF method that uses a number of hand-crafted features. 
	The third row in Table \ref{table:essay} gives the performance of
	Bi-LSTM-CRF with ground-truth argumentative-or-not information
	for each sentence, i.e. the feature $p$ in Figure 3 are ground-truth labels;
	surprisingly, this method even outperforms the ``human upperbound'' 
	performance\footnote{The human upperbound performance is obtained by 
		by averaging the evaluation scores of all three annotator pairs on test data.
		Note that sentences' argumentative-or-not information are not used
		in obtaining the human upperbound performance.
	} reported in \cite{stab2016parsing}, validating our assumption
	that the sentences' argumentative-or-not information is helpful
	for the ACBD task. This is validated again by the outperformance
	of our joint model against Bi-LSTM-CRF (row 5 and 6 in Table \ref{table:essay}). 
	Note that, among all methods that do not use the ground-truth
	argumentative-or-not information, our joint model achieves the highest 
	performance.
	
	%As a second benchmark, we present results on the 
    The performances on the
	Wikipedia articles are presented in Table \ref{table:classification} and 
    Table \ref{table:boundary}. 
    The upper part of these two tables give the performances
    of some existing ACBD methods, and we can see that 
    the performance metrics used for existing methods
    and for our RNN-based methods are different:
    our RNN-based methods output a unique
    boundary and component type for the input sentence, thus the performance
    metric is P/R/F1; however, existing ACBD methods
    produce a ranked list of candidate argument component
    boundries, thus their performance metrics are, e.g.,
    precision@200, i.e. the probability that the 
    true boundary is included in the top 200 predicted boundaries
    (definitions of recall@200 and F1@200 can be obtained similarly).
    Also note that,
	the results reported in \cite{levy2014context} are obtained from 
	a slightly older version of the dataset, containing only 32 topics 
	(instead of 33) and 976 claims (instead of 1332). 
	
    From Table \ref{table:classification}, we find that
	for argumentative sentence classification, our joint model 
	significantly outperforms all the other techniques.
    From Table \ref{table:boundary},
	we find that the joint RNN model prevails over 
    the other Bi-LSTM based models, again confirms that
    the argumentative-or-not 
	information can further improve the boundary detection performance. 
    Note that, performances on
	Wikipedia corpus are not that high in general.
	One of the reasons is that the length of the argument 
    component is long and the performance metrics we use are
    strict. In addition, only topic-dependent claims 
	are annotated in the Wikipedia corpus;
	our RNN-based approaches do not consider the
	topic information, thus identify some topic-irrelevant 
	claims, which are treated as False Positive.
	Similar observations are also made in \cite{lippi2015context}.
	
	\section{Conclusion}
    In this work, we present the first deep-learning based
    family of algorithms for the argument component boundary
    detection (ACBD) task. In particular, we propose a novel
    \emph{joint model} that combines an attention-based
    classification RNN to predict the argumentative-or-not
    information and a Bi-LSTM-CRF network to identify the 
    exact boundary. We empirically compare the joint model with 
    Bi-LSTM, Bi-LSTM-CRF and some state-of-the-art 
    ACBD methods on two benchmark corpora;
    results suggest that    
    our joint model outperforms all the other methods, 
    suggesting that our joint model can effectively 
    use the argumentative-or-not information 
    to improve the boundary detection performance. 
    As for the future work, a natural next step 
    is to apply deep learning
    techniques to other sub-tasks of argumentation mining;
    in addition, a deep-learning-based end-to-end argumentation 
    mining tool is also worthy of further investigation.


	
	
	%\bibliographystyle{IEEEtran}
	
	\bibliography{boundary_detection}
	\bibliographystyle{plain}
	
\end{document}